\documentclass{article}
\usepackage{spconf,amsmath,graphicx}

\usepackage{booktabs} 
\newcommand*{\thead}[1]{\multicolumn{1}{c}{\bfseries #1}}
\makeatletter \providecommand{\@LN}[2]{} \makeatother
\usepackage{xcolor}

\usepackage{amssymb}

\DeclareMathAlphabet{\pazocal}{OMS}{zplm}{m}{n} 
\usepackage{mathtools}
\DeclarePairedDelimiterX{\norm}[1]{\lVert}{\rVert}{#1}

\title{MetricBERT: Text Representation Learning via Self-Supervised Triplet Training}
\name{Itzik Malkiel$^{1,2}$\thanks{$^{*}$ Denotes equal contribution.}$^{*}$~~~Dvir Ginzburg$^{1,2,*}$~~~Oren Barkan$^{1,3}$~~~Avi Caciularu$^{4}$~~~Yoni Weill$^{1}$~~~Noam Koenigstein$^{1,2}$
}
\address{$^1$Microsoft, $^2$Tel-Aviv University, $^3$The Open University, $^4$Bar-Ilan University}

\begin{document}

\maketitle
\begin{abstract}
We present MetricBERT, a BERT-based model that learns to embed text under a well-defined similarity metric while simultaneously adhering to the ``traditional'' masked-language task. We focus on downstream tasks of learning similarities for recommendations where we show that MetricBERT outperforms state-of-the-art alternatives, sometimes by a substantial margin. 
We conduct extensive evaluations of our method and its different variants, showing that our training objective is highly beneficial over a traditional contrastive loss, a standard cosine similarity objective, and six other baselines. As an additional contribution, we publish a dataset of video games descriptions along with a test set of similarity annotations crafted by a domain expert\footnote{https://zenodo.org/record/6088355}. 
\end{abstract}
%
%
\section{Introduction}
\label{sec:intro}
Learning textual similarity is a long-standing task with applications in information retrieval \cite{yang2019simple}
, document clustering \cite{thakur2021beir}
, essay scoring \cite{ke2019automated}
, and recommender systems \cite{malkiel-etal-2020-recobert,CB2CF,barkan2021cold,barkan2021coldicdm,barkan2020attentive,barkan2020cold,nam}. 
In this work, we propose MetricBERT, a model for text similarity that employs a novel metric learning technique. 
MetricBERT is a self-supervised model that utilizes 
pseudo-labels of negative and positive text passages extracted from the dataset at hand.
The pseudo-labels serve as a proxy to the ultimate (but absent) similarity labels. 
While MetricBERT is a general model for text similarity, we chose to focus on text similarity for recommender systems where we show that MetricBERT outperforms the current state-of-the-art by a sizeable margin.

MetricBERT jointly optimizes a masked-language model and a novel margin-based triplet (metric learning) loss. The metric learning strives to embed similar samples closer to each other, compared to dissimilar ones as explained in Sec.~\ref{sec:method}. Additionally, MetricBERT leverages a hard negative mining procedure~\cite{dong2017class} to accelerate convergence and improve performance. 

In this work, we provide an extensive comparison between different models, focusing on downstream tasks of learning similarities for recommendations. In particular, we report the results of seven methods - LDA, TF-IDF, BERT~\cite{bert}, RoBERTa~\cite{roberta}, RecoBERT~\cite{malkiel-etal-2020-recobert}, SBERT~\cite{sbert}, and our proposed MetricBERT, evaluated on two datasets for wines and video games recommendations. Notably, we show that the choice of the triplet loss over a traditional contrastive loss is highly beneficial and that optimizing the angular distance, over the alternative of cosine similarity, is crucial for model convergence. Finally, as an additional contribution, we publish a new dataset for text-based similarity of video games along with a test set of similarity annotations crafted by a domain expert.

\begin{figure*}[t]
\includegraphics[width=0.79\linewidth]{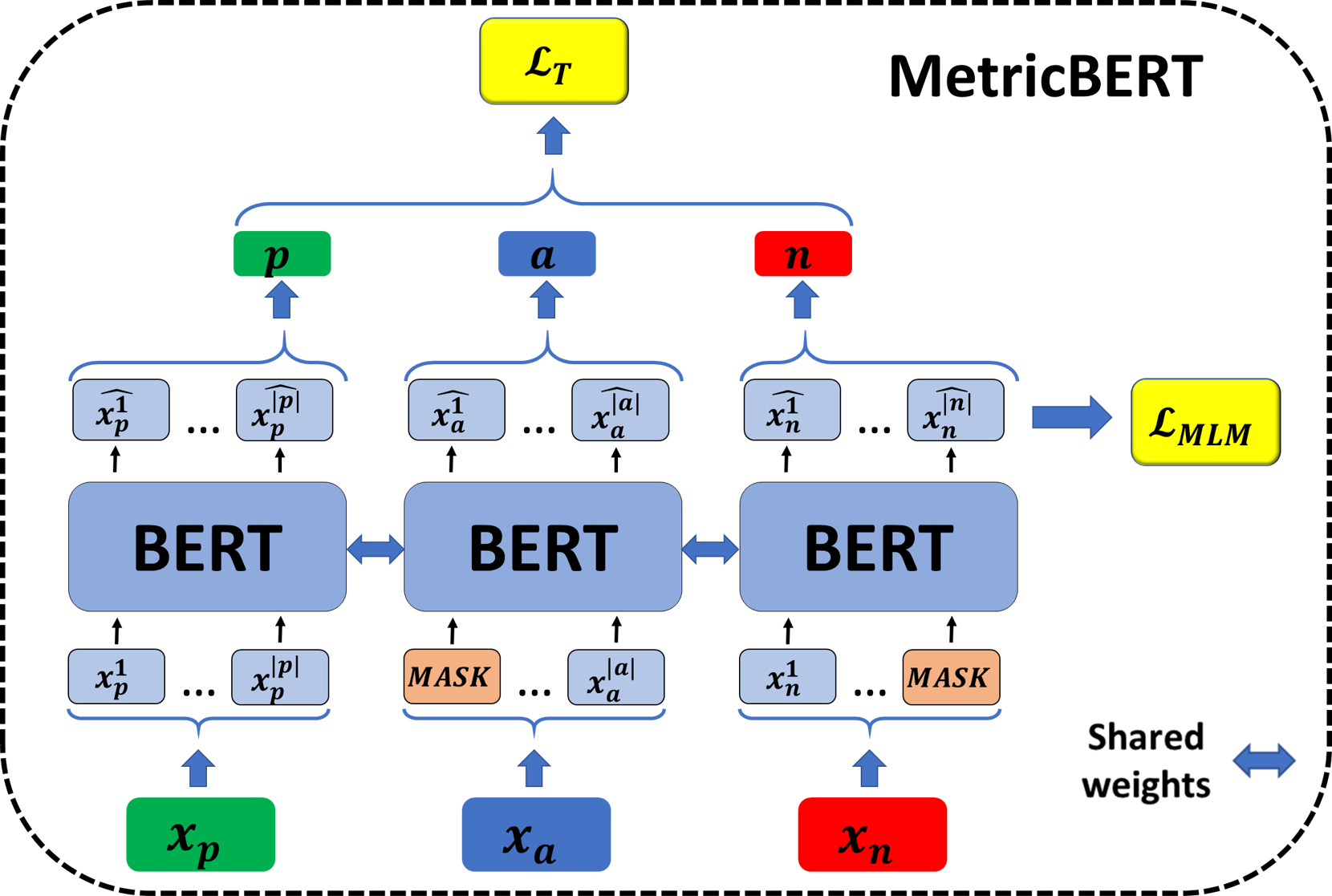}
\centering
\caption{Schematic illustration of MetricBERT. Triplets of textual elements are propagated through the model, which separately embeds their tokens. The embedded tokens, of each element, are (1) fed into a masked language model loss and (2) pooled into a single vector. The pooled three vectors are applied through the triplet loss objective $\pazocal{L}_{T}$.
}
    \label{fig:metricBERT}
\end{figure*}

\section{Related Work}

Recent Transformer-based representation learning methods in NLP rely on training large neural language models (LMs) on self-supervised objectives and massive corpora~\cite{bert,roberta}
. These models consists of a two-phase training procedure: First the LM is trained on large unlabeled corpora. Then, the resulting pre-trained model is trained with supervision for a specific downstream task. These models show great promise in various linguistic tasks
\cite{bert,ginzburg-etal-2021-self,barkan2020scalable, roberta, malkiel-wolf-2021-maximal, blaier-etal-2021-caption}
, and usually produce better results than non-contextualized neural embedding methods~\cite{mikolov2013distributed,barkan2017bayesian,barkan-etal-2020-bayesian,vbn,barkan2016item2vec,barkan2020explainable,barkan2021anchor,barkan2016modelling} .

Recent works suggested new approaches for improving the sentence embedding abilities of transformer-based representation learning models: For example, SBERT~\cite{sbert} proposed using distantly-supervised natural language inference (NLI) to create representations that significantly outperform other state-of-the-art sentence embedding methods.
In the domain of text-based recommendations, RecoBERT~\cite{malkiel-etal-2020-recobert} proposed a BERT-based model for text-based item recommendations, by continuing the training of the pre-trained BERT model on the catalog at hand. RecoBERT utilizes BERT's original pre-training objective of masked language modeling, along with a novel title-description objective that leverages an additional self-supervision task to match between the embeddings of an item title and its description.
In this work, we focus on a textual recommendation task and compare (among other models) to both \cite{sbert} and \cite{malkiel-etal-2020-recobert}.

Another relevant line of work investigated integrating metric learning techniques for LMs, mostly for the scenario where supervised data is abundant. SPECTER~\cite{cohan-etal-2020-specter} was introduced for producing document-level representations for scientific documents. 
SPECTER embeds scientific documents by pre-training a Transformer language model using supervision obtained by citation graphs. 
Due to the lack of similarity labels (or citations) within our domain, we were unable to compare our method with SPECTER.
Recently, in \cite{rethmeier2021primer} the authors surveyed contrastive learning concepts for NLP applications and their relations to other fields, and concludes with an open call to bring more contrastive methods to NLP.

\section{Method}
\label{sec:method}

Let $D := \left\{(a_i,b_i)^{N}_{i=1}\right\}$ be a dataset of $N$ items, where each item $i$ is a pair of two textual elements. 
Given a source item $s \in D$, the task is to rank all the other items in $D$ according to their semantic similarity to $s$.

MetricBERT training is initialized with a pre-trained BERT model $f$. 
The training proceeds using triplets of textual elements $(x_a, x_p, x_n)$ where $x_a$, $x_p$, and $x_n$ are \emph{anchor}, \emph{positive} and \emph{negative} samples, respectively.
In our settings, each anchor and positive samples are textual elements associated with the same item $p \in D$, while the negative sample is a different, randomly sampled item $n \in D$ according to the procedure described in Sec.\ref{sec:mining}.

The textual elements are separately tokenized and masked (similar to BERT pre-training), aggregated over the batch dimension, and propagated through $f$. 
The embeddings of each of the three elements are used to train MetricBERT to (1) rank the positive and negative samples, based on the anchor text, and (2) reconstruct the masked words in each element. This is obtained by minimizing a combination of a novel triplet-loss, and a standard masked language loss\footnote{similar to BERT} denoted by $\pazocal{L}_{T}$ and $\pazocal{L}_{MLM}$, respectively. 
The triplet loss ($\pazocal{L}_{T}$) is defined by:
\begin{equation}
\label{eq:triplet-loss}
\pazocal{L}_{T}  = L(a, p, n) =  \max (0, m+d(a, p)-d(a, n)),
\end{equation}
where $m$ is a pre-defined margin, $a, p, n$ are the embeddings: 
\begin{equation}
a =f\left(x_{a}\right), \; \; p =f\left(x_{p}\right), \; \; n =f\left(x_{n}\right),
\end{equation}
and $d$ is the angular distance, which can be expressed as:
\begin{equation}
d(u, v)=
\arccos \left(
C(u, v)
\right) 
/ \pi
\end{equation}
where $u,v \in \mathbb{R}^{d}$ are d-dimensional vectors and $C$ is the cosine similarity:
\begin{equation}
C(u, v) = \frac{u \cdot v}{\|u\|\|v\|}
\end{equation}

The choice of the angular distance over a standard cosine similarity stems from early empirical observations that optimizing a cosine similarity results in a sizeable degradation in performance, where the model “collapses” into a narrow manifold (i.e. the embedding of all items in the dataset retrieved a cosine similarity that approaches 1). 
This improved performance of the angular distance can be attributed to its enhanced sensitivity to micro-differences in the angle between vectors with similar directions. 

The improved sensitivity of the angular distance is also apparent in the derivatives of the two metrics. While the derivative of the cosine distance is: 
\begin{equation}
\frac{\partial}{\partial u} \operatorname{C}(u, v)=\frac{v}{\|u\| \|v\|}-\operatorname{C}(u, v) \cdot \frac{u}{\|u\|^{2}}
\end{equation}
the angular distance derivative can be expressed as:
\begin{equation}
\frac{\partial}{\partial u} d(u, v)= -\frac{1}{\pi \sqrt{1-\left(\operatorname{C}(u, v)\right)^{2}}} \frac{\partial}{\partial u} \operatorname{C}(u, v)
\end{equation}
for which, compared to the cosine derivative, the denominator dynamically scales the gradients by an inverse ratio of the cosine distance between the vectors. In other words, compared to the cosine similarity, the angular distance yields gradients of larger magnitude for vectors with similar directions. See Sec.~\ref{sec:ablation} for more details.

Finally, the total loss of MetricBERT is:
\begin{equation}\label{eq:tot_loss}
\pazocal{L}_{total} =\pazocal{L}_{MLM} + \lambda \pazocal{L}_{T},
\end{equation}
where 
$\lambda$ is a balancing hyper-parameter. The MetricBERT model is illustrated in Fig.~\ref{fig:metricBERT}.

\subsection{Mining Hard Negatives}
\label{sec:mining}
MetricBERT constructs triplets by associating each anchor-positive pair with the hardest negative sample in a given batch. More specifically, given an anchor-positive pair, the angular distance between the anchor and all the other elements in the batch is calculated (neglecting the ``positive'' element associated with the anchor). Next, the element with the smallest distance from the anchor is selected as the negative sample. This technique allows the optimization to focus on the miss-matched elements that ``confuse'' the model the most, retrieve more triplets that violate the margin, and enhance the gradients of the triplet loss term.

\subsection{Inference}
\label{subsec:inference}
Given a source and candidate items $s := (a_i,b_i), c:= (a_j,b_j)$, we score their similarity by: $\text{Metric}\textsubscript{Inf}(s,c) = d(a_i,a_j) + d(b_i,b_j)$.
Item ranking is then retrieved by sorting, in ascending order, all candidate items according to their similarity with $s$.

\begin{table*}[t!]
\centering
\resizebox{0.85\linewidth}{!}{%
\begin{tabular}{@{}l@{~~}c@{~~}c@{~~}c@{~~}c@{~~}c@{~~}c@{~~}c@{~~}c@{~~}c} 
\toprule
&& \multicolumn{4}{c}{Video games}& \multicolumn{4}{c}{Wines}\\
\cmidrule(l){3-6}
\cmidrule(l){7-10}
\textbf{Model}& \textbf{Inference}&  \textbf{
MPR\ \ \ \ }&  \textbf{MRR}&
\textbf{HR@10}& \textbf{HR@100}&  \textbf{MPR\ \ }&  \textbf{MRR}&
\textbf{HR@10}& \textbf{HR@100}\\
\midrule

LDA& \multicolumn{1}{c}{-} 
&79.8\%&4.5\%&1.9\%&8.6\%
&85.2\%&24.9\%&9.7\%&52.2\%
\\
tf-idf& \multicolumn{1}{c}{-} 
&80.3\%&20.0\%&8.7\%&22.8\%
&96.1\%&85.4\%&61.8\%&89.2\%\\

  \midrule 
BERT&  Mean
&83.3\%&18.2\%&5.0\%&21.3\%  
&87.5\%&58.7\%&22.4\%&65.3\% \\
BERT& Metric\textsubscript{Inf} 
&87.3\%&21.7\%&7.8\%&24.9\%  
&92.8\%&70.3\%&31.6\%&75.2\% \\

  \midrule
    
RoBERTa& Mean 
&73.3\%&22.2\%&5.8\%&14.2\%  
&86.2\%&62.6\%&24.9\%&64.3\%\\
RoBERTa& Metric\textsubscript{Inf} 
&74.9\%&23.6\%&6.1\%&16.2\%  
&93.6\%&84.0\%&42.6\%&76.6\%\\

  \midrule
  
RecoBERT & RecoBERT\textsubscript{Inf} 
&84.1\%&24.5\%&8.5\%&25.3\%  
&96.3\%&91.7\%&65.4\%&94.9\% \\

 SBERT& \multicolumn{1}{c}{-} 
&84.5\%&20.3\%&8.7\%&21.9\%
&93.3\%&77.9\%&43.2\%&79.9\% \\

  \midrule 
  
MetricBERT& Metric\textsubscript{Inf} 
&\textbf{89.1\%}&\textbf{25.0\%}&\textbf{9.1}\%&\textbf{26.0}\% 
&\textbf{98.5\%}&\textbf{95.3\%}&\textbf{74.7\%}&\textbf{95.6\%}  \\
 
\bottomrule
\end{tabular}}

\caption{Similarity results evaluated on the video games (left), and wines (right) datasets based on expert annotations. 
The second column specifies the applied inference method, where "Mean" refers to mean-pooling the token-embeddings. 
}
\label{Tab:results}
\end{table*}

\section{Experimental Setup and Results}
We evaluate MetricBERT on the publicly available wines recommendations dataset\footnote{https://doi.org/10.5281/zenodo.3653403} as well as the video-games recommendations dataset\footnote{https://zenodo.org/record/6088355}, which we release as an additional contribution, to further accelerate research in the field. 

\subsection{The Datasets}
The Wines dataset~\cite{malkiel-etal-2020-recobert} contains $\sim$130K wines, each associated with a name (single sentence describing the wine type, grape variety, year, etc.) and a review written by a professional wine reviewer. 
Performance is evaluated on a test set of similarity labels annotated by a professional sommelier.

The Video Games dataset consists of 21,935 video game articles excerpted from Wikipedia. The ``Gameplay'' section was taken from each article, resulting in pairs of game names and text passages describing the game characteristics. 
As an additional contribution, we publish a test set of 90 source articles, each associated with $\sim 10$ similarity annotations crafted by an expert.

\subsection{The Baselines}
We compare MetricBERT to the following baselines:

\noindent{{\bf LDA and TF-IDF\quad}} Latent Dirichlet Allocation (LDA) and  Term Frequency–Inverse Document Frequency (TF-IDF) are renowned axiomatic methods for topic modeling and text-matching. As LDA is extremely sensitive to hyper-parameter tuning we followed the procedure from~\cite{ginzburg-etal-2021-self} that perform a grid search on the highest topic coherence. For the evaluations of TF-IDF, we computed the cosine similarity between the obtained TF-IDF vectors

\noindent{{\bf RecoBERT\quad}}
this is a self-supervised BERT-based model trained to predict item similarities~\cite{malkiel-etal-2020-recobert}. RecoBERT optimizes the MLM loss and a title-description model that relies on ``negative'' and ``positive'' pairs of text. The latter model utilizes a classification head on top of the cosine representations of the elements. The RecoBERT inference (RecoBERT\textsubscript{Inf}) ranks the items by utilizing the scores obtained from the trained classification head and the cosine between the pooled embedding of the textual elements.
 
\noindent{{\bf BERT and RoBERTa\quad}}
are the BERT$_{\text{LARGE}}$ \cite{bert} and the RoBERTa$_{\text{LARGE}}$\cite{roberta} models, which continued the pre-training on each of the given datasets.

\noindent{{\bf SBERT\quad}}
this model utilizes a training approach that produces semantically meaningful embeddings under a cosine-similarity metric~\cite{sbert}. Different from RecoBERT, SBERT requires supervision to optimize its embedding under the cosine metric. Since similarity labels are not available, we adopt the publicly available SBERT model, without continuing training on the given dataset. 

\subsection{Evaluation metrics}
Performance is evaluated 
by utilizing the Mean Percentile Rank (MPR), Mean Reciprocal Rank (MRR), and Hit Ratio at $k$ (HR@$k$) metrics, which are widely used for the similarity and ranking tasks \cite{malkiel-etal-2020-recobert, ginzburg-etal-2021-self}.

%
\subsection{Results}

\label{sec:results}
The results, depicted in Tab.~\ref{Tab:results}, are based on the ground-truth expert annotations associated with each dataset. 
For all the MetricBERT experiments, we set $\lambda = 1$ in Eq.~\ref{eq:tot_loss}. 
As can be seen, MetricBERT outperforms all other models by a sizeable margin. 
Notably, in the wines similarity task, MetricBERT improves MPR by 2.2, MRR by 3.6, and HR@10 by 9.3 w.r.t the best performing model.

Compared to axiomatic methods (LDA and TF-IDF), MetricBERT achieves better results in all metrics on both datasets. We attribute this to the ability of a well-trained transformer model to produce accurate contextualized embeddings. As to the BERT and RoBERTA models, the results in Tab.~\ref{Tab:results} indicates the contribution of MetricBERT's training strategy.

Interestingly, 
applying Metric\textsubscript{Inf} inference to other baseline language models improves performance by a sizeable margin. This is another evidence for applicability of the proposed inference Metric\textsubscript{Inf}, which separately pools the embeddings of each of the textual elements, before inferring the angular distance. 

SBERT, as opposed to MetricBERT, 
only offers a \emph{supervised} training procedure for new datasets. Tab.~\ref{Tab:results} shows that MetricBERT outperforms SBERT on these datasets by 4.6\% on MPR and 4.7\% on MRR for the video games dataset and yields even larger gaps on the wines dataset.

\begin{table*}[t]

\centering
\begin{tabular}{@{}l@{~~}l@{~~}c@{~~}c@{~~}c@{~~}c} 

&\thead{Model} & \thead{MPR} & \thead{MRR} & \thead{HR@10} & \thead{HR@100} \\
\hline


(i)& Random Mining&   {97.5\%}  &  {93.1\%} & {67.6\%} & {94.1\%} \\

(ii)&Contrastive Loss&   {96.3\%}  &  {86.3\%} & {61.2\%}&
{89.8\%}\\
(iii)&Cosine Loss&     {96.3\%}  &  {86.3\%} & {61.2\%}&
{89.8\%} \\

(iv)&Cosine Similarity &  {96.1\%}  &  {90.5\%} & {69.3\%} &
{90.7\%}\\
(v)&No MLM &  {89.1\%}  &  {81.5\%} & {45.3\%} &
{79.7\%}\\

&\textbf{MetricBERT}&   \textbf{98.5\%}&
\textbf{95.3\%}&
\textbf{74.7\%}&
\textbf{95.6\%} \\
\hline

\hline
\end{tabular}
\caption{Ablation study results. The differences between the full method and all other variants are statistically significant with $p < 0.05$.}
\label{Tab:ablation}
\end{table*}

\subsection{Ablation Study}
\label{sec:ablation}
Table~\ref{Tab:ablation} presents an ablation study for MetricBERT, evaluated on the wines dataset. 
The following variants are considered: (i) Random Mining - choosing negative samples in each batch randomly (as opposed to the hard mining procedure in Sec.~\ref{sec:mining}.
(ii, iii) Contrastive/Cosine loss - trained on pairs instead of triplets by replacing Eq.~\ref{eq:triplet-loss} with: $\mathcal{L}_{cont} = [m_{pos} - s_p]_+ + [s_n - m_{neg}]_+$, where $m_{pos},m_{neg}$ are the positive and negative margins respectively, and $s_p,s_n$ are the positive and negative similarity values between the embeddings under the cosine metric. For the contrastive evaluation we set $m_{pos}=1, m_{neg}=0$ while for the cosine evaluation $m_{pos}=1, m_{neg}=-1$. (iv) Cosine Similarity - replace the angular distance with a cosine similarity metric. (v) No MLM - Optimize only the similarity objective without the MLM loss.

The results, shown in Tab.\ref{Tab:ablation}, showcase the significant contribution of the triplet loss as well as that of our hard mining strategy. Specifically, compared to both the contrastive loss (ii) and the cosine loss (iii) variants, we see improvements of 2\% MPR and 9\% MRR. 
We attribute these substantial improvements to the dynamics of the triplet loss, which do not penalize positive pairs if they are close enough to each other w.r.t the negative element. This is in contrast to both the contrastive and the cosine losses, which persistently attempt to force maximal similarity of 1 between positive text pairs, at the expense of the LM objective. 

Note that in this experiment, the results of the contrastive loss (ii) and cosine loss (iii) are identical. 
This stems from the fact that during training,
no negative pair had a negative cosine value, thus in this experiment, both variants had identical gradient steps throughout the entire training procedure. In the general case, these two variants may yield results that are slightly different from each other, however still inferior to the MetricBERT model proposed in this paper. 

As can be seen in Tab.~\ref{Tab:ablation}(iv), substituting the angular distance with a raw cosine similarity yields a sizeable degradation in performance, where the model “collapses”, and embeds the textual elements in a narrow manifold (retrieving a cosine similarity that is very close to 1 for all element-pairs in the dataset). We observe that this collapse mode is attributed to the natural properties of the cosine loss, which is not sensitive to embeddings with similar angles, compared to the angular distance which 
produces gradients with a relatively larger magnitude for embeddings with similar directions.

\section{Discussion and Conclusions}
We introduce MetricBERT, a BERT-based model that jointly optimizes a triplet-loss and a masked language model. 
We focus on text-based recommendation tasks where we demonstrate the ability of MetricBERT to improve upon the state-of-the-art sometimes by a considerable margin.
While this work is focused on a recommendation task, MetricBERT is a general model, that can be utilized for additional text similarity and information retrieval tasks.
Finally, to accelerate research and as an additional contribution, we publish a dataset of video games descriptions along with a test set of similarity annotations crafted by a domain expert.

\bibliographystyle{IEEEbib}
\bibliography{main,anthology}

\end{document}